\documentclass[10pt,twocolumn,letterpaper]{article}

\usepackage[pagenumbers]{cvpr} % To force page numbers, e.g. for an arXiv version

\definecolor{cvprblue}{rgb}{0.21,0.49,0.74}
\usepackage[numbers]{natbib} 
\usepackage[pagebackref,breaklinks,colorlinks,allcolors=cvprblue]{hyperref}
\usepackage{wrapfig}
\usepackage{amsmath}
\usepackage{amssymb}
\usepackage{bbm}

\newcommand{\specialcell}[2][c]{%
  \begin{tabular}[#1]{@{}c@{}}#2\end{tabular}}

\newcommand\blfootnote[1]{%
  \begingroup
  \renewcommand\thefootnote{}\footnote{#1}%
  \addtocounter{footnote}{-1}%
  \endgroup
}

\usepackage{multirow}

\def\paperID{*****} % *** Enter the Paper ID here
\def\confName{CVPR}
\def\confYear{2026}

\title{Breakdance Video classification in the age of Generative AI}

\author{Sauptik Dhar, Naveen Ramakrishnan, Michelle Munson\\
Eluvio AI Labs\\
918 Parker Street, Berkeley, CA 94710 \\
{\tt\small sauptik.dhar@eluv.io, naveen.buckeye@gmail.com, michelle.munson@eluv.io}
}

\begin{document}
\maketitle
\begin{abstract}
 Large Vision Language models have seen huge application in several sports use-cases recently. Most of these works have been targeted towards a limited subset of popular sports like soccer, cricket, basketball etc; focusing on generative tasks like visual question answering, highlight generation. This work analyzes the applicability of the modern video foundation models (both encoder and decoder) for a very niche but hugely popular dance sports - breakdance. Our results show that Video Encoder models continue to outperform state-of-the-art Video Language Models for prediction tasks. We provide insights on how to choose the encoder model and provide a thorough analysis into the workings of a finetuned decoder model for breakdance video classification.       
\end{abstract}    
\section{Introduction}

\label{sec:intro}
The advent of Vision Language models have seen an emergence of their application in several domains including social media \cite{zeng2024large}, search and recommendation \cite{wu2024surveylargelanguagemodels, search_git}, healthcare \cite{he2025surveylargelanguagemodels, liu2024surveymedicallargelanguage}, AI assistants for cognitive tasks \cite{luo2025largelanguagemodelagent, agent_git} etc \blfootnote{\copyright\;   Eluvio AI Labs. Naveen Ramakrishnan contributed in an individual capacity}. Although these systems have been successfully integrated in production, very little focus has been placed on sports; particularly accurate identification and natural language description of the sports actions. Ubiquitous to most sports use-cases is the accurate categorization of the sport-centric actions. Sports activity recognition is crucial to several use cases like, content discovery (retrieval of the top moments in game play \cite{xu2025deeplearningsportsvideo}) , content understanding (captioning or summarization of the game clips \cite{shih2017survey, dhar2025largevlmbasedstylizedsports, electronics14030461}), and even content generation (5, 10, 15 - min highlight generation of a game \cite{calagari2017sports, midoglu2024ai}). All these tasks need proper domain-centric tagging of the game play. Although there have been recent efforts in leveraging video foundation models for accurate activity recognition in sports; such works are limited to a few popular sports like, soccer, cricket, football, basketball etc. \cite{xu2025deeplearningsportsvideo, shih2017survey, xia2024languagemultimodalmodelssports}. Moreover, there are no studies on the comparison of recent state-of-the-art (Encoder-based) Video Foundational Models \cite{vivit, vidmae, v_jepa} vs. (Decoder-based) Video Language models \cite{qwen, gemma3n} for activity recognition. A good comparison study especially for niche  sports' action classification would lend significantly to future application of the more appropriate advanced approach for several other uncharted sports.  

\begin{figure*}
  \begin{center}
    \includegraphics[width=\textwidth]{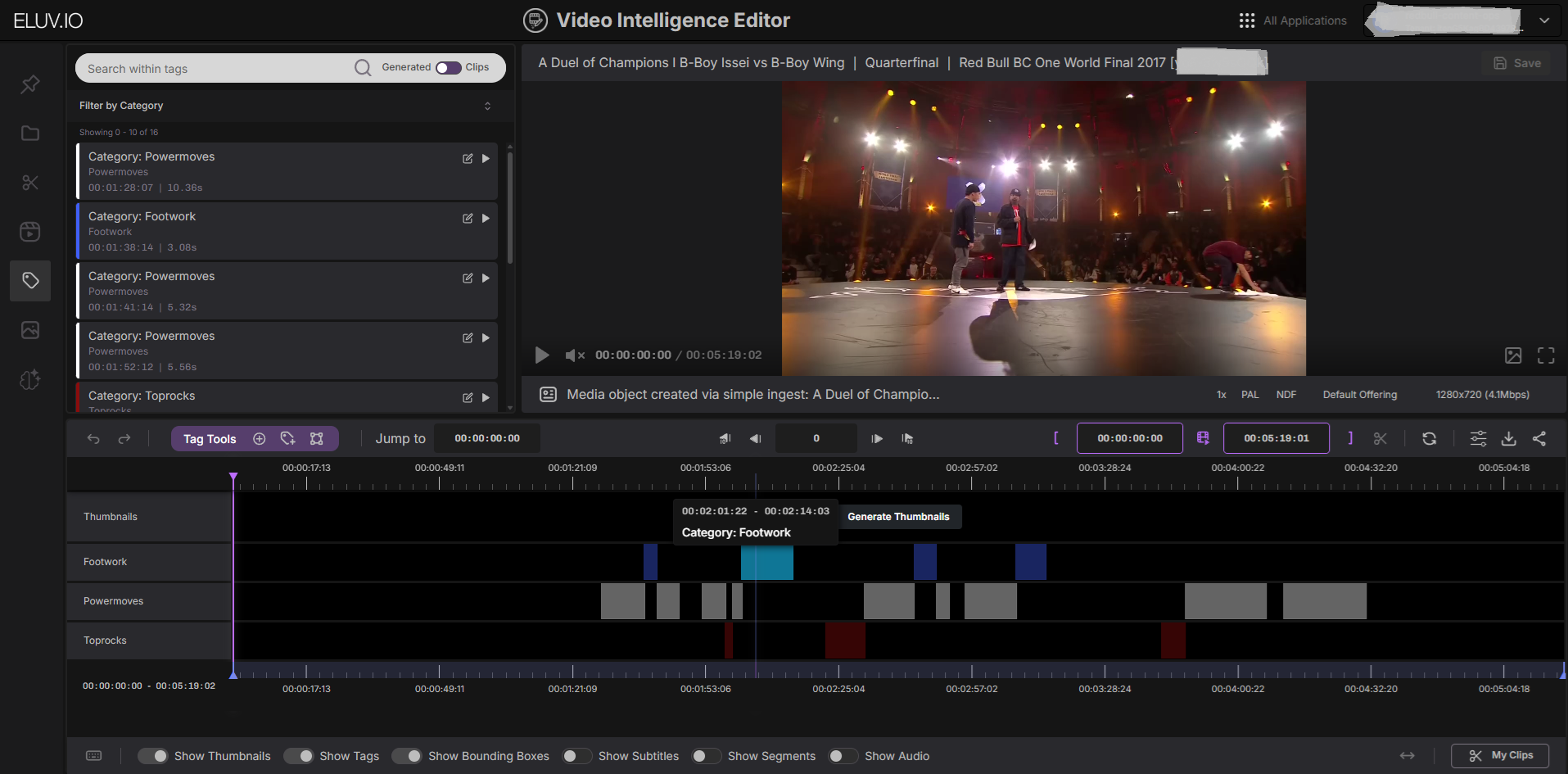}
    \end{center} 
  \caption{Breakdance Video and time stamped tags in Eluvio Video Intelligence Editor (EVIE)} \label{fig:evie} 
\end{figure*} 

This work compares the performance of the state-of-the art (Encoder-based) Video Foundational Models \cite{vivit} \cite{vidmae} \cite{imagebind} and (Decoder-based) Video Language models \cite{qwen} for a niche dance sports -  Break dance \cite{breakdance1, breakdance2}. Break dancing has seen a huge surge in the dance sports community and was the first dance sport to compete at the Olympics (Paris 2024). Also break dance sport has similarities with other high-speed, intermittent-style sports (e.g. basketball, tennis) in terms of the spatio-temporal structure of the data; so, we believe our analysis and insights based on the break dance sport could be applicable to other sports. We analyze the strength and shortcomings of state-of-the-art video (encoder vs decoder) models for downstream video activity classification. The main contributions of this work are as follows:
\begin{enumerate}
    \item We extend several state-of-the-art video foundation (encoder-based) models and the recent Qwen-2.5-VL (decoder-based) model for break-dance move classification, and provide a thorough comparative analysis between these approaches.  
    \item We provide several ablation studies highlighting the advantages and disadvantages of the models and derive insights into the workings of the encoder embeddings as well as the finetuned decoder models.
    \item All our codes and the processed data (embeddings) are publicly released to promote open research. 
\end{enumerate}
To the best of our knowledge, this is the first work providing a thorough analysis on -
\begin{itemize}
    \item[--] Comparison of Encoder vs Decoder based video foundation models for \textbf{action classification}. 
    \item[--] Analyzing the state-of-the-art for \textbf{video-level} activity classification of a rising dance sport - \textbf{breakdance}.
\end{itemize}

% \begin{figure*}
%   \centering
%   \begin{subfigure}{0.68\linewidth}
%     \fbox{\rule{0pt}{2in} \rule{.9\linewidth}{0pt}}
%     \caption{An example of a subfigure.}
%     \label{fig:short-a}
%   \end{subfigure}
%   \hfill
%   \begin{subfigure}{0.28\linewidth}
%     \fbox{\rule{0pt}{2in} \rule{.9\linewidth}{0pt}}
%     \caption{Another example of a subfigure.}
%     \label{fig:short-b}
%   \end{subfigure}
%   \caption{Example of a short caption, which should be centered.}
%   \label{fig:short}
% \end{figure*}

% \input{sec/2_rel_works}
\section{BRACE: The Breakdancing Competition Dataset}
\label{sec:problem}

    The BRACE dataset \cite{brace} consists of videos from the Red Bull BC One breakdancing competition. The data features the best dancers in the world competing against each other. The competition follows a knockout tournament format, where two dancers compete in a 1-vs-1 battle taking turns to perform a number of breaking sequences. A good summary of the sequences in this dataset is provided in Table \ref{tab:brace}.

    \begin{table}
    \centering
    \begin{small}
    \caption{Summary of BRACE dataset \cite{brace}} \label{tab:brace}
    \begin{tabular}{l|l}
        \hline
        % \addlinespace
         Frames (Annotated) & 334538 (26676) \\
         Sequences & 465 \\
         Segments & 1352 \\
         Duration & 3h 32m\\
         Dancers & 64\\
         Videos & 81\\
         Avg. segments per sequence & 2.91\\
         Avg. sequence duration & 27.48s\\
         Avg. segment duration & 9.45s\\
         \hline
    \end{tabular}
    \end{small}
    \end{table}

    \begin{figure}
      \begin{center}
        \includegraphics[width=0.45\textwidth]{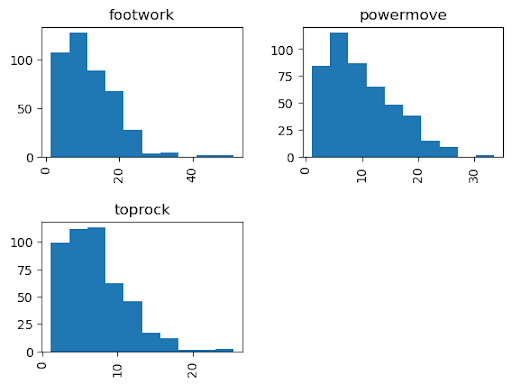}
        \end{center} 
      \caption{The label distribution. x - axis move duration in seconds, y - axis frequency of video segments. The move durations (in secs) for a) footwork = 11.5 $\pm$ 6.95 sec, b) powermove = 9.85 $\pm$ 5.9 sec, c) toprock = 7.01 $\pm$ 4.01 sec.} \label{label_freq} 
    \end{figure}

\begin{figure*}
      \begin{center}
        \includegraphics[width=\textwidth]{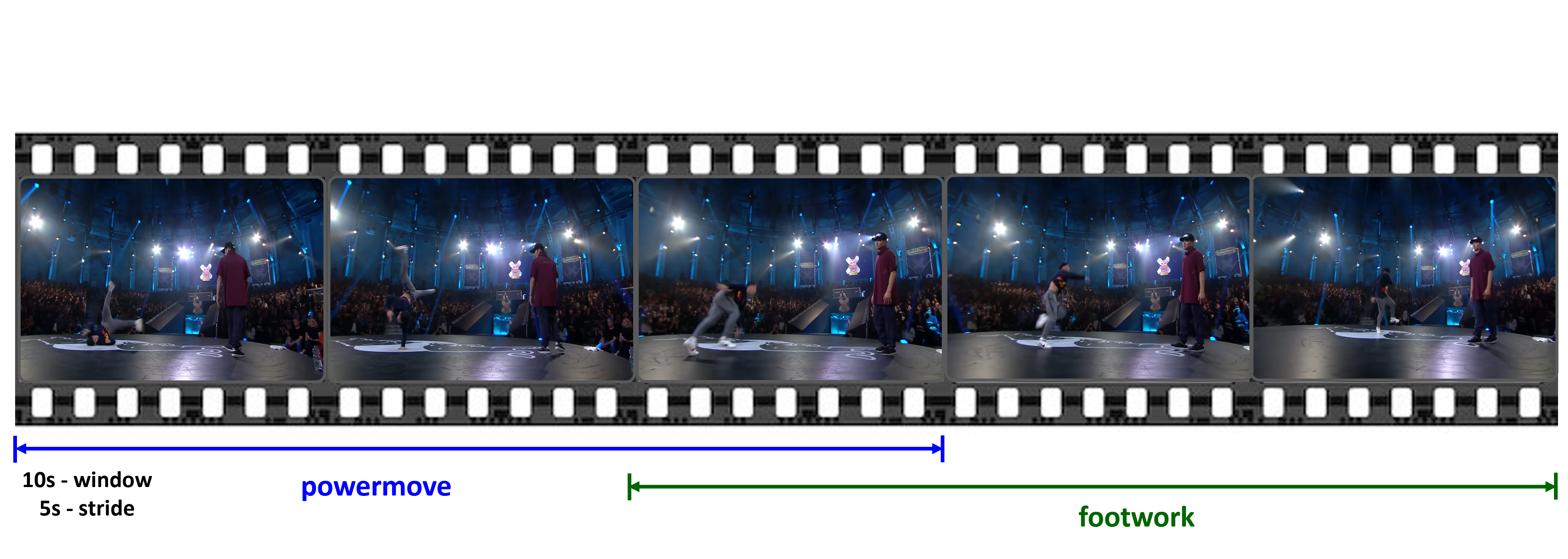}
        \end{center} 
      \caption{Schematic representation of the breakdance video's moving window prediction.} \label{fig:break_reel} 
    \end{figure*}

    For this work the task is to recognize the different dance moves from a video stream. There are three dance moves labeled in this data. These are,
    \begin{itemize}
        \item[--] \textbf{Powermoves} \cite{breakdance1, breakdance2} are dynamic and acrobatic movements in breakdance. They have an element of continuous rotation across one or multiple body parts, for example spinning on the head, hand or back. Powermoves are often perceived as the most impressive part of the dance, as they can look highly impressive. 
        \item[--] \textbf{Footwork} \cite{breakdance1, breakdance2} is a fundamental element of breakdancing where the dancer moves their legs in intricate patterns on the floor while their hands provide support. It involves using the hands and feet to create rhythmic, circular, or linear movements across the floor and is also known as ``downrock" or ``floorwork".
        \item[--] \textbf{Toprock} \cite{breakdance1, breakdance2} is the standing, upright dance style performed in breaking (breakdancing), which serves as an entry to more advanced floor moves. It involves rhythmic footwork and arm movements that allow dancers to showcase their individual style, coordination, and rhythm while dancing to the music.
    \end{itemize}
    The label distribution is provided in Fig \ref{label_freq}. A typical breakdance video from the Brace data with its respective time-stamped move tags is shown in Fig. \ref{fig:evie}. Here we illustrate using the Eluvio Video Intelligence editor (EVIE) \cite{evie}. The goal is to predict if a given video segment has either of the moves (`powermove', `footwork', `toprock'). For a segment with no moves we label as `None'. Further, we predict a dance move within a window-segment of 10 sec and an overlapping stride of 5 sec. A high-level schema of this approach is shown in Fig. \ref{fig:break_reel}. This 10s window size is chosen based on the average move duration as shown in Fig \ref{label_freq}. Increasing the window segments further leads to significant overlapping of the moves and is not desirable; whereas a very small window segment could lead to less context to predict the move. For this data set, increasing or decreasing the window segments did not provide any significant improvement in performance.

% https://breakdancingninja.com/Music/Encyclopedia_Of_Breakdancing.pdf
% https://www.mygrooveguide.com/dance-info/breaking/power-moves

\section{Video Foundation Models} \label{sec:methods}
We explore several models for the move classification. These models can broadly be categorized as Encoder or Decoder - based models.

\subsection{Encoder - Models}
We extend the encoder-based video foundation models for breakdance move classification. Modern encoder based approaches process video inputs by first tokenizing video patches (spatial and temporal) and passing them through a transformer architecture. Typically a special token (\texttt{CLS}) embedding is passed through a classification head. Traditionally, an encoder-based approach is preferred for predictive tasks like video classification, segmentation etc.  

\begin{figure}
\vspace{-1.2cm}
  \begin{center}
    \includegraphics[width=0.4\textwidth]{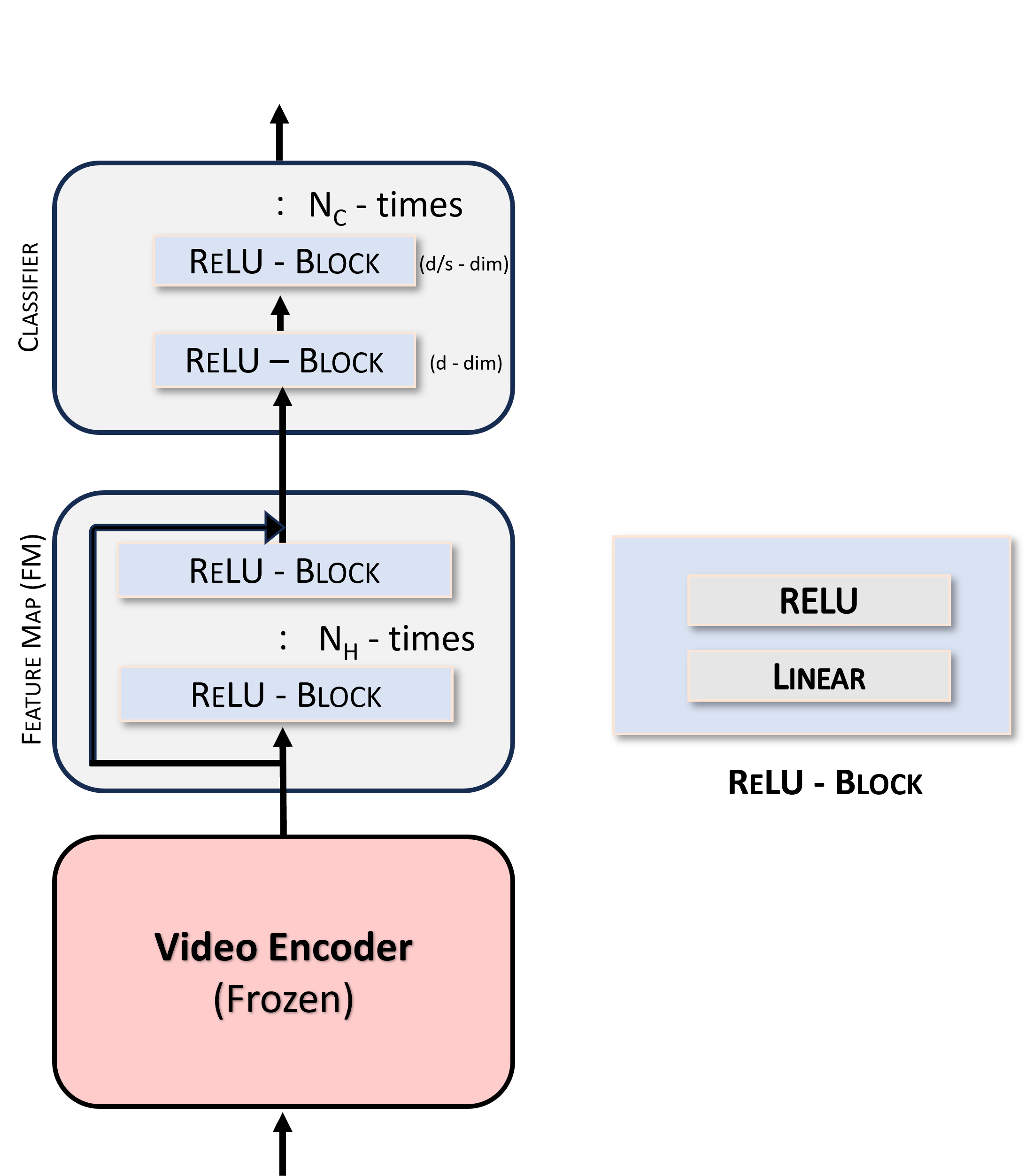}
  \end{center} 
  \caption{Encoder Model Architecture.} \label{fig_encoder_model} 
  \vspace{-0.6cm}
\end{figure}

\begin{figure}
  \begin{center}
    \includegraphics[width=0.30\textwidth]{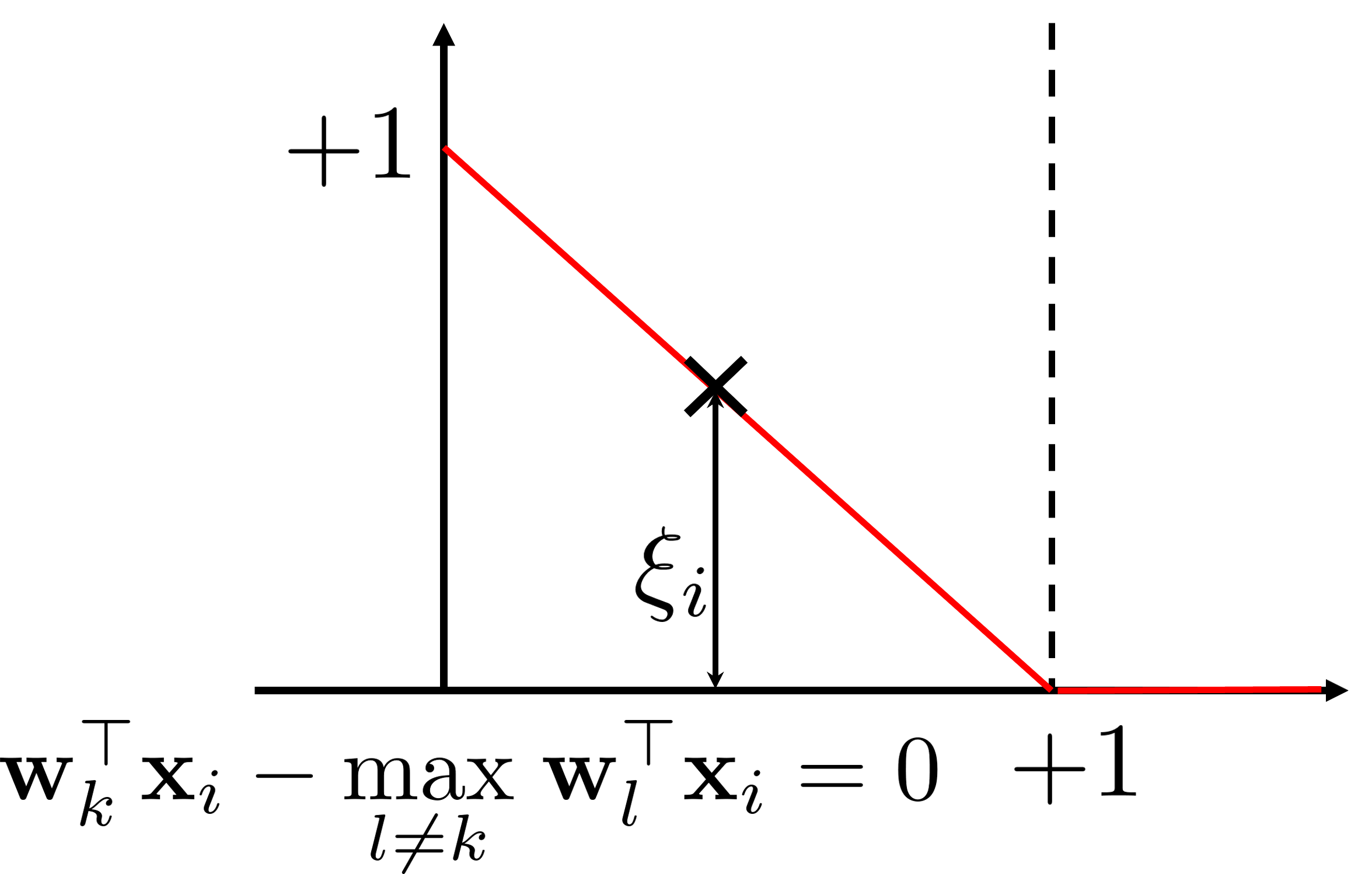}
  \end{center} 
  \caption{Multiclass Hinge loss for linear mapping $f(\mathbf{x}) = \mathbf{w}^\top \mathbf{x}$. Sample $(\mathbf{x}_i,y_i)$ lying inside the margin is linearly penalized using slack variable $\xi_i$.} \label{fig_hinge} 
\end{figure}

\begin{figure}
  \begin{center}
  \vspace{-0.2cm}
    \includegraphics[width=0.35\textwidth]{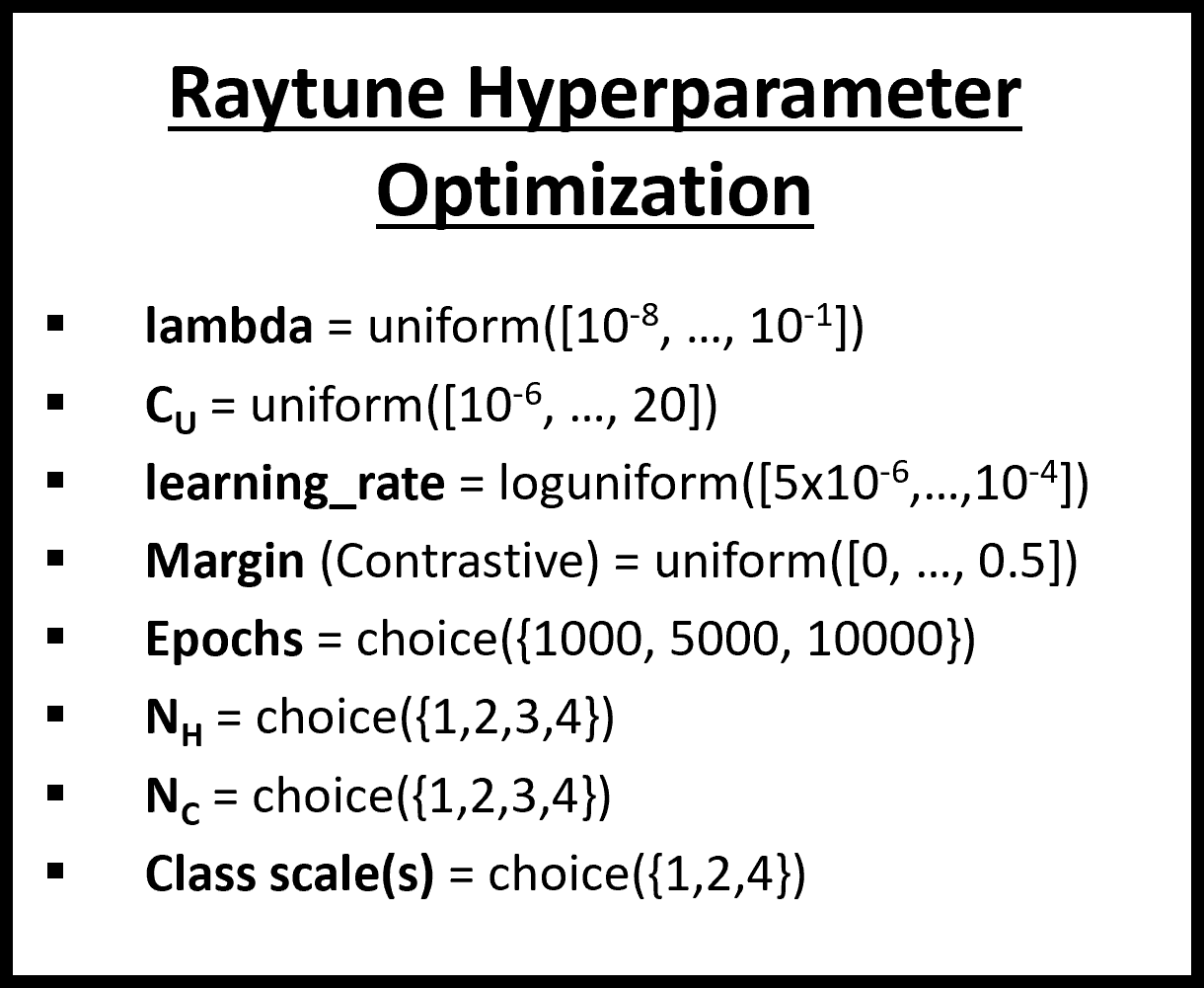}
  \end{center} 
  \caption{Encoder Model hyperparameter optimization using Raytune \textsc{Asha} scheduler.} \label{fig_hpo} 
\end{figure}

For our representative encoder-based approach, we use a pretrained video encoder, and add FCN - \textsc{ReLU} blocks at its end (see Fig \ref{fig_encoder_model}). These additional blocks can be categorized into two parts. The Feature Map (FM) block, which performs nonlinear mapping from a $d$ - dimensional vector to another $d$ - dimensional vector. We create $N_H$ copies of this block. The optimal $N_H$ is selected through hyperparameter optimization (discussed in section \ref{sec:exp}). We use a residual connection across all the hidden nodes. The second part is a classifier network, which maps the output of FM - block to $L$ - dimensional logits, where $L = 4$ is the number of classes. The number of hidden layers is parameterized by $N_C$. The dimension of each layer is scaled by a factor of $s$. For training we use a contrastive loss at the output of the FM - block,

\begin{align} 
\label{eq_cos_embed_loss}
 L_C^{ij} = \; \left\{
\begin{array}{l l}
   1- \text{cos}(\mathbf{x}_i, \mathbf{x}_j) & \text{if } y_{ij} = 1 \\
   (\text{cos}(\mathbf{x}_i, \mathbf{x}_j) - \Delta)_+ &  \text{else}
\end{array}\right. \ 
\end{align}
where, $(\cdot)_+$ = relu operator, and $\Delta$ = margin. And a multiclass hinge loss \eqref{eq_cs_loss} (see Fig \ref{fig_hinge}) at the classifier head.
The multiclass hinge loss adopts a multi-valued function $\mathbf{f} = [f_1, \ldots, f_L]$ for the decision rule, 
% \begin{align} 
% \label{eq_dec_rule}
%  h(\mathbf{x}) \; \left\{
% \begin{array}{l l}
%    =  k & \text{if } f_k(\mathbf{x}) > f_\ell(\mathbf{x}) \;;  \forall \ell \neq k  \\
%    \neq [1,\ldots, L] &  \text{else}
% \end{array}\right. \ 
% \end{align}

\begin{align} \label{eq_dec_rule}
h(\mathbf{x}) = \underset{k}{\text{argmax}} \; f_k(\mathbf{x}) &&
\end{align}

\noindent and minimizes the loss, 
\begin{flalign} 
\label{eq_cs_loss}
& L_T = \sum\limits_{i=1}^n  \max_{k \in \mathcal{Y}} \; \{ 1-\delta_{ik} + f_k(\mathbf{x}_i)-f_{y_i}(\mathbf{x}_i)\} &&
\end{flalign}
where, $\delta_{i\ell}=\mathbbm{1}_{(y_i=\ell)}$. The final loss function is, 
\begin{flalign} 
\label{eq_tot_loss}
 &\underset{\textbf{W}}{\text{min}} \quad \lambda||\textbf{W}||^2 \; + L_T +  C_U \sum\limits_{ij} L_C^{ij} &&
\end{flalign}
\noindent During training we freeze the weights of the Video Encoder and only update the FM and Classifier network. The model hyperparameters during training is tuned using the Raytune framework with the \textsc{Asha} - scheduler \cite{raytune}. Details on the raytune hyperparameter setup are provided in Fig \ref{fig_hpo}.

For our experiments we use several state-of-the-art encoder-based video foundation models. We use, 
\begin{itemize}
    \item[--] \textbf{Video Vision Transformer - ViViT} \cite{vivit} is a milestone architecture that extends the success of the Vision Transformer to videos. The model aims to extract the spatio-temporal tokens (called tubelets) from input videos and encodes them by a series of transformer layers. The ViVit architecture dictates that the input video be represented as 32 frames. Inline to the design we uniformly sample 32 frames for each 10 sec video window segments. We use the \texttt{google/vivit-b-16x2-kinetics400} pretrained weights throughout this paper.
    \item[--] \textbf{Video Masked Autoencoder - Video MAE} \cite{vidmae} is a self-supervised learning framework for training powerful video representation models. Inspired by the success of Masked Autoencoders (MAE) for image representation learning, Video MAE extends this approach to video data, enabling efficient and data-efficient training of video Transformers. The Video MAE architecture dictates that the input video be represented as 16 frames. Inline to the design we uniformly sample 16 frames for each 10 second video window segments. We use the \texttt{MCG-NJU/videomae-base-finetuned-kinetics} pretrained weights throughout this paper.
    \item[--] \textbf{ImageBind} \cite{imagebind} is a multimodal model that binds several modalities including audio, video, text, depth, thermal, IMU with images, and provides a shared embedding space for all modalities. The shared embedding space captures the semantic relation across these modalities. Imagebind dictates that every 2 second video segment is represented by 2 frames. To accommodate the high variability of motion at high frame-rate for break dancing, here for each 5 sec non overlapping window we use a 20 equally spaced overlapping moving window of 2 seconds. Each of these 2 seconds moving windows are represented using 2 frames. We use the pretrained \texttt{imagebind-huge} throughout the paper.   
\end{itemize}

\subsection{Decoder - Models}

\begin{figure*}
  \begin{center}
    \includegraphics[width=\textwidth]{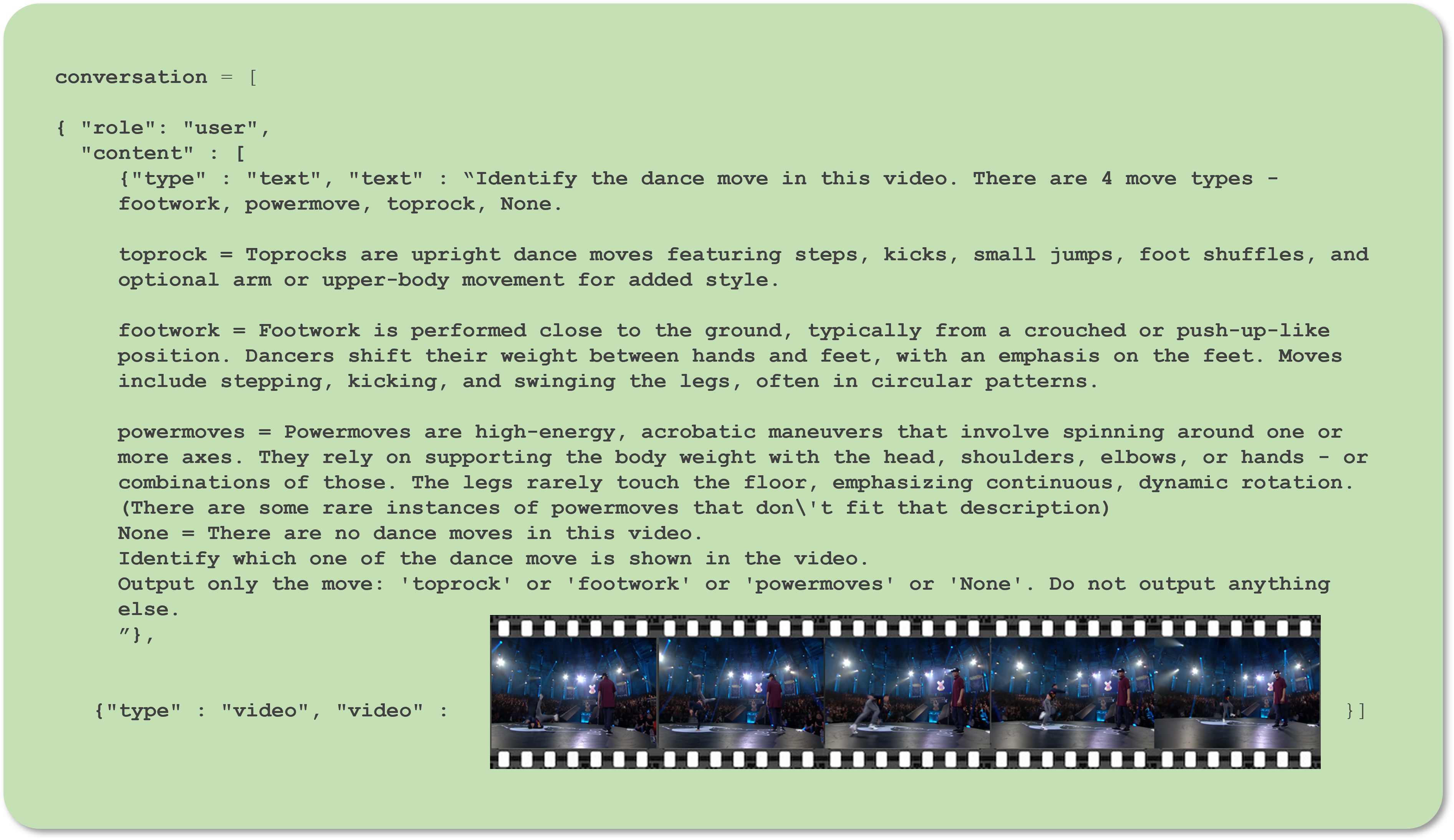}
  \end{center} 
  \caption{Zero shot prompt format} \label{fig_zs_prompt} 
\end{figure*} 

\begin{figure*}
  \begin{center}
    \includegraphics[width=\textwidth]{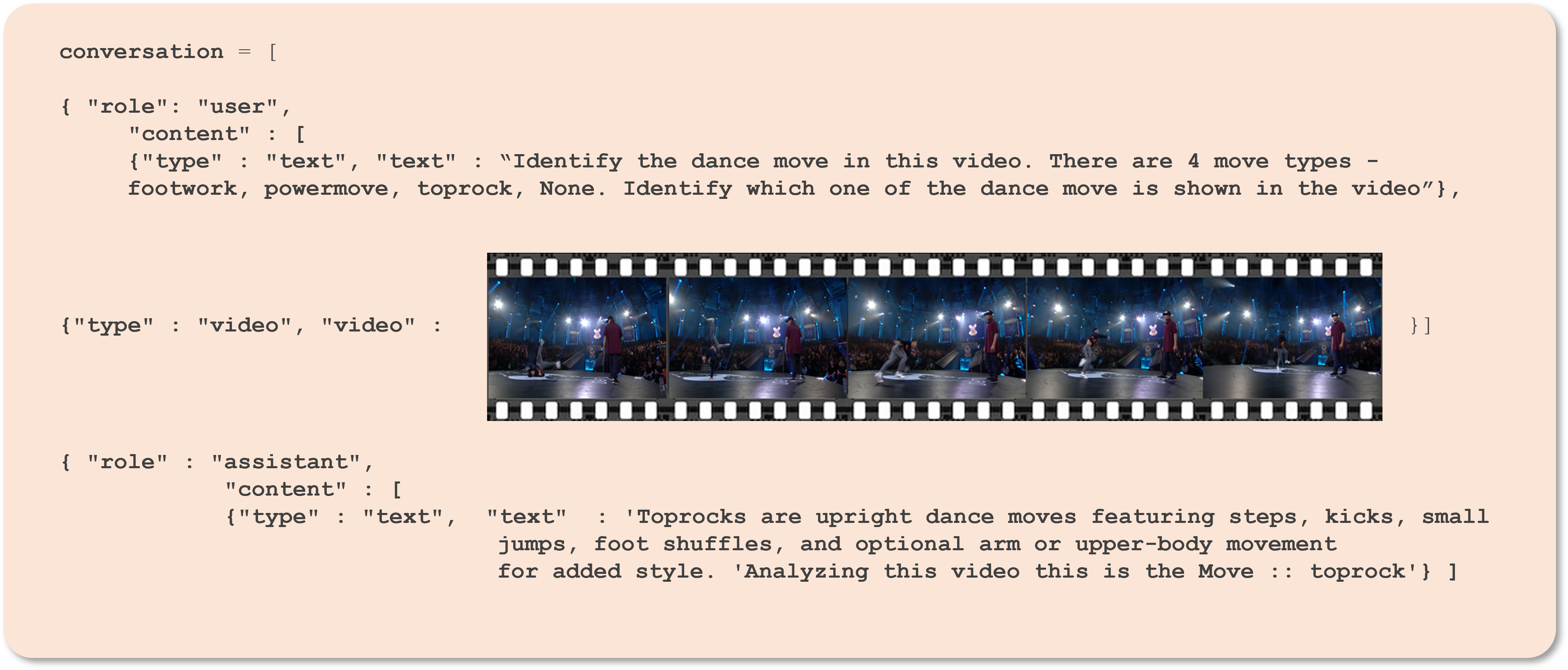}
  \end{center} 
  \caption{Finetuning prompt format} \label{fig_ft_prompt} 
\end{figure*} 

\begin{figure}
  \begin{center}
  \vspace{-0.2cm}
    \includegraphics[width=0.35\textwidth]{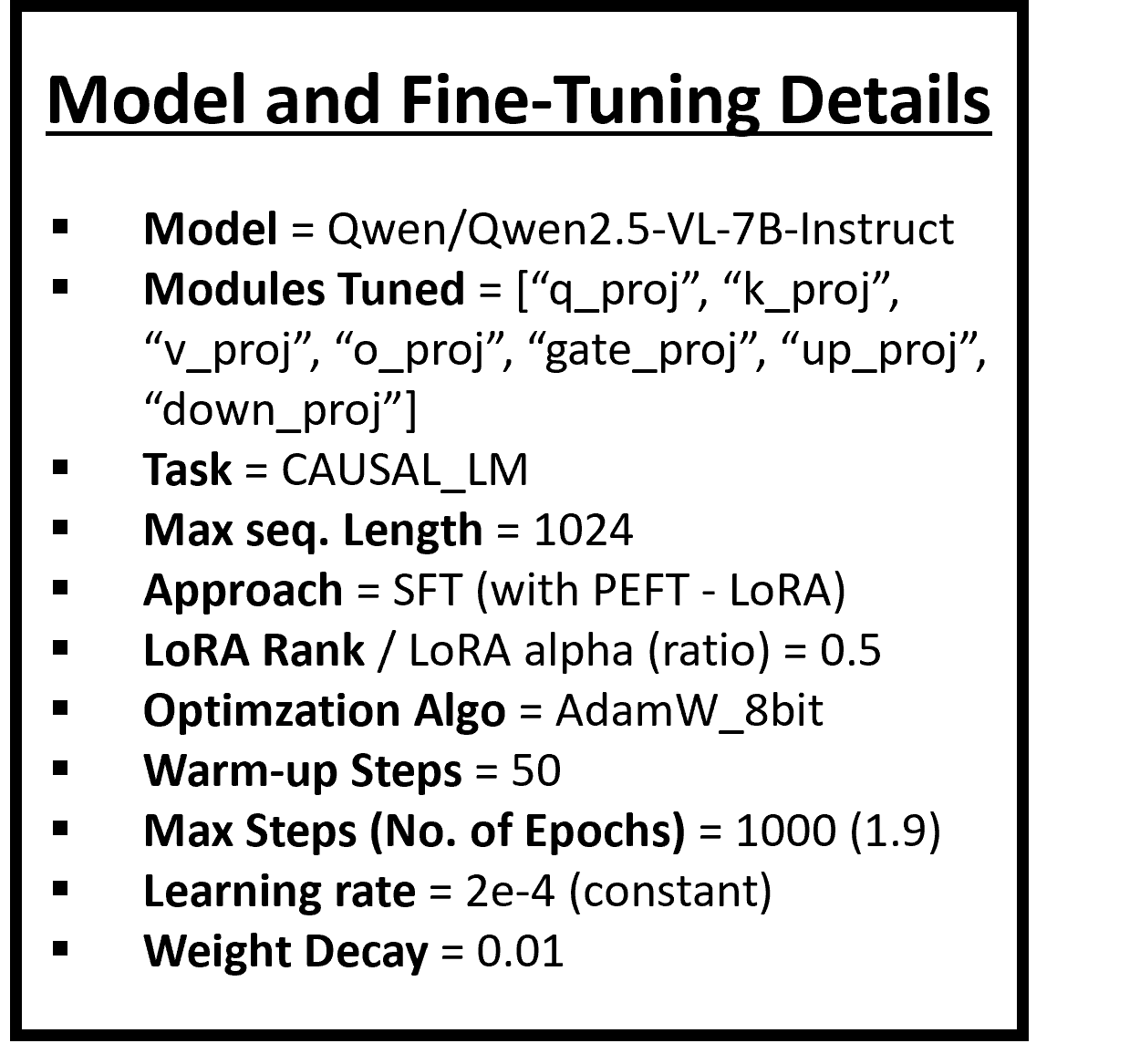}
  \end{center} 
  \caption{Instruction finetuning parameters for \texttt{Qwen2.5-VL}.} \label{fig_finetune_hpo} 
\end{figure}

As an alternate approach we leverage the modern decoder-based Video Large Language Model (VLM) to identify the moves. Such systems use a video processing component (a.k.a. ``video decoder") to interpret visual information from videos and integrate it with the language model for analysis, conversation, and generation. These systems enable an LLM to understand a video's content, actions, and temporal dynamics and respond in a human-like way. For our work we use the Qwen2.5-VL \cite{qwen} as a representative of this approach. The instruction and the prompt template for the different versions are provided in Fig \ref{fig_zs_prompt} (zero-shot) and Fig \ref{fig_ft_prompt} (fine-tuning). Here for a comparable analysis with the encoder based approaches we try two different frame rates and represent 10s video segments with 16 and 32 frames i.e. 1.6 fps and 3.2 fps respectively. Details on the instruction fine-tuning parameters are provided in Fig \ref{fig_finetune_hpo}.   

    % instruction = '''Identify the dance move in this video. There are 4 move types - footwork, powermove, toprock, None. 
    % toprock = Toprocks are upright dance moves featuring steps, kicks, small jumps, foot shuffles, and optional arm or upper-body movement for added style.
    % footwork = Footwork is performed close to the ground, typically from a crouched or push-up-like position. Dancers shift their weight between hands and feet, with an emphasis on the feet. Moves include stepping, kicking, and swinging the legs, often in circular patterns.
    % powermoves = Powermoves are high-energy, acrobatic maneuvers that involve spinning around one or more axes. They rely on supporting the body weight with the head, shoulders, elbows, or hands - or combinations of those. The legs rarely touch the floor, emphasizing continuous, dynamic rotation. (There are some rare instances of powermoves that don\'t fit that description)
    % None = There are no dance moves in this video.
    % Identify which one of the dance move is shown in the video. 
    % Output only the move: 'toprock' or 'footwork' or 'powermoves' or 'None'. Do not output anything else.
    % '''
    
    % conversation = [
    %     { "role": "user",
    %       "content" : [
    %         {"type" : "text",  "text"  : instruction},
    %         {"type" : "video", "video" : frames} ]
    %     }

\section{Experiment and Results}\label{sec:exp}

\begin{table*}[!h]
\centering
\begin{small}
\caption{Frame Level Accuracy for the best model.}
\tabcolsep=0.4cm
\begin{tabular}{c|c*{4}{|c}}
\hline
\hline
\multicolumn{2}{c|}{\multirow{2}{*}{Method}} & \multicolumn{2}{c|}{Test} & \multicolumn{2}{c}{Train} \\
\cline{3-6}
\multicolumn{2}{c|}{} & Overall & Per Video & Overall & Per Video\\
\hline
\multirow{3}{*}{\centering \rotatebox[origin=c]{0}{\specialcell{\textsc{Encoder}}}} & \textsc{ViVit} & $0.59$ & $0.59 \pm 0.08$ & $0.69$ & $0.64 \pm 0.07$\\
% \cline{2-5}
& \textsc{VidMAE} & $0.66$ & $0.62 \pm 0.07$ & $0.7$ & $0.64 \pm 0.07$\\
% \cline{2-5}
& \textsc{ImageBind} & $\mathbf{0.69}$ & $\mathbf{0.66 \pm 0.08}$ & $\mathbf{0.75}$ & $\mathbf{0.69 \pm 0.08}$\\
\hline
\multirow{3}{*}{ \centering  \rotatebox[origin=c]{0}{\specialcell{ \textsc{Decoder}\\(Zero-shot)}}} & \specialcell{\textsc{Qwen 2.5VL} (3B)} & $0.28$ & $0.28 \pm 0.05$ & $0.26$ & $0.26 \pm 0.02$ \\
% \cline{2-5}
& \specialcell{\textsc{Qwen 2.5VL} (7B)} & $0.47$& $0.46 \pm (0.1)$ & $0.46$ & $0.45 \pm 0.08$  \\
& \specialcell{\textsc{Qwen 2.5VL} (32B)} & $0.54$& $0.53 \pm (0.07)$ & $0.52$ & $0.5 \pm 0.07$ \\
\cline{1-6}
\multirow{2}{*}{ \centering  \rotatebox[origin=c]{0}{\specialcell{ \textsc{Decoder} \\ (Finetuned)}}} & \specialcell{\textsc{Qwen 2.5VL} (7B)-1.6fps} & $0.62$ & $0.59 \pm 0.09$ & $0.71$ & $0.67 \pm 0.06$ \\
& \specialcell{\textsc{Qwen 2.5VL} (7B) - 3.2fps} & $0.62$ & $0.59 \pm 0.07$ & $0.66$ & $0.62 \pm 0.07$ \\
\hline
\hline
\end{tabular}

\label{tab:main}
\end{small}
\end{table*}

\subsection{Encoder vs Decoder - Generalization}

For our first (and main) experiment we compare the performance of the encoder vs decoder approaches for video classification. For this we partition the data as Training  = 71 titles and Test = 10 titles. Additional details on data partitioning (test videos' ids) are provided in Appendix \ref{sec:app_brace_details}). 

Table \ref{tab:main} provides the performance of the different approaches. Only the result of the best performing model after hyperparameter tuning is reported.   Here we provide the accuracy at the frame level. In addition we also provide the per video frame accuracy. We report the optimal encoder model's performance using \textsc{Asha} hyperparameter optimization. For the Decoder based model we provide the results for zero-shot for different model sizes \texttt{Qwen2.5-VL-Instruct} (3B, 7B and 32B) model. In addition we also provide the results for the instruction-tuned model using \texttt{Qwen2.5-VL-7B-Instruct} and (4bit quantized - nf4 representation).   

As seen from the results in Table \ref{tab:main} the encoder-based models provide superior performance compared to the decoder based ones. While the state-of-the-art decoders may enjoy significant advantage due to (almost always) large scale pretraining, the finetuning is mostly catered towards token generation, and may provide inferior performance on prediction task. This is inline to the current understanding of the advantages of encoder vs decoders. While lately there have been a few research illustrating superior gains using decoder based LLMs for text classification \cite{textllm}, this does not yet hold for video level classification.  

\subsection{Encoder Embeddings}
\begin{figure}
  \centering
  \begin{subfigure}{0.4\linewidth}
    \includegraphics[width=1.2\textwidth]{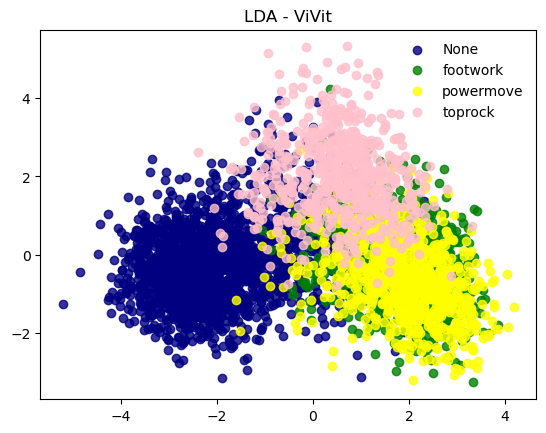}
    \caption{LDA - ViVit}
    \label{fig:short-a}
  \end{subfigure}
  \hfill
  \begin{subfigure}{0.4\linewidth}
    \includegraphics[width=1.2\textwidth]{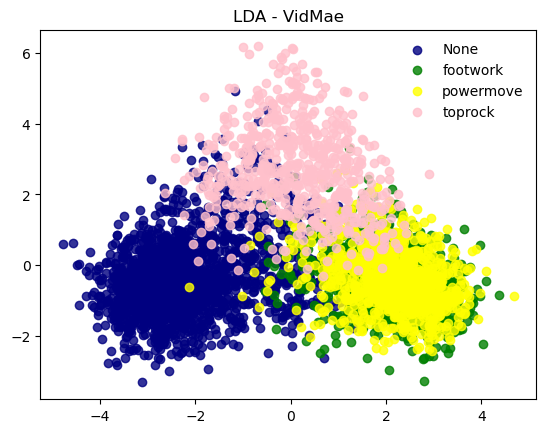}
    \caption{LDA - VidMae}
    \label{fig:short-b}
  \end{subfigure}
  \hfill
 \begin{subfigure}{0.4\linewidth}
    \includegraphics[width=1.2\textwidth]{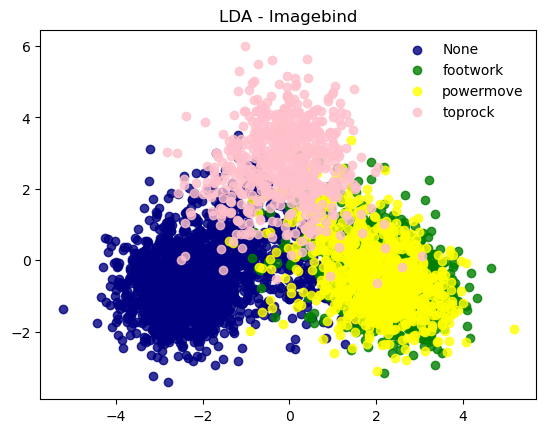}
    \caption{LDA - Imagebind}
    \label{fig:short-a}
    \end{subfigure}
  \caption{LDA analysis on the embedding space induced by the video encoders.}
  \label{fig:short}
\end{figure}

\begin{table}[!h]
\centering
\begin{small}
\caption{LDA - separability score. Higher is better.}
\tabcolsep=0.4cm
\begin{tabular}{c|c|c}
\hline
Methods & $\mathbf{J_1}$ & $\mathbf{J_2}$ \\
\hline
ViVit & $3.20$ & $2.56$ \\
VidMAE & $3.52$ & $2.96$ \\
Imagebind & $3.92$ & $3.24$ \\
\hline
\end{tabular}
\label{tab:lda}
\end{small}
\end{table}

Next we dig deeper into the performance of the Encoder based models. As seen from the results in Table \ref{tab:main} the imagebind embeddings provided better generalization. To explore a bit more we analyze the embedding space for the different embedding models. For this we perform a Fisher Linear Discriminant Analysis (LDA) on the embedding spaces induced by the different pretrained frozen video models. We further measure the separability score for the different classes along the first two LDA loading vectors i.e. we measure, $J(\mathbf{w}) = \frac{\mathbf{w}^\top S_B \mathbf{w}}{\mathbf{w}^\top S_W \mathbf{w}}$; where $S_B$ = between class and $S_W$ = within class scatter matrix, and evaluate it along the first-two LDA directions, 
\begin{align}\label{eq_ldascores}
    &  J_1 = J(\mathbf{w}_1) \quad \text{where} \; \mathbf{w}_1 = \underset{\mathbf{w}}{\text{argmax}} \; J(\mathbf{w}) \nonumber && \\
    & J_2 = J(\mathbf{w}_2) \quad \text{where} \; \mathbf{w}_2 = \underset{\mathbf{w} \neq \mathbf{w}_1}{\text{argmax}} \; J(\mathbf{w}) \nonumber\nonumber &&
\end{align}

\noindent Table \ref{tab:lda} provides a comparison of the separability score for the different embeddings. The results shows that the ImageBind embedding provides highest separability followed by VidMAE and ViViT. Note that there are subtle differences on how the embedding space had been pretrained. While ViViT and VidMAE target classifying the kinetics - 400 classes, Imagebind is designed to capture the overall semantics across modalities - images, text, videos etc. It is likely that such a space capture the generic semantics over modalities and can better adapt to new tasks. This is also seen in Table \ref{tab:main} where Imagebind outperforms both VidMAE and ViViT. Note however, applicability of this general rule of using a linear operation (like LDA) as a relative measure for task generalization needs to be confirmed through additional analysis over several datasets. However, it does provide a simple explanation of Imagebind's superior performance relative to the other approaches.

\subsection{Decoder Finetuning - Ablation Study}

\begin{table*}
\centering
\begin{footnotesize}
\caption{Frame Level Accuracy ($\times 100$) for different \texttt{Qwen2.5-VL-7B-Instruct} instruction tuning setups. Minimum stopping criteria set to \texttt{avg token accuracy}$>97\%$.}
\tabcolsep=0.2cm
\begin{tabular}{c|c|cccc|cccc}
\hline
\hline
\multicolumn{2}{c|}{\multirow{3}{*}{\textsc{Rank}}}& \multicolumn{4}{c|}{\textsc{Greedy Decode}} & \multicolumn{4}{c}{\specialcell{\textsc{Not Greedy}\\(\textsc{Temperature} = 1.0)}} \\
\cline{3-10}
\multicolumn{2}{c|} {}& \multicolumn{2}{c}{\textsc{No Desc.}} & \multicolumn{2}{c|}{\textsc{Add Desc.}}& \multicolumn{2}{c}{\textsc{No Desc.}}  & \multicolumn{2}{c}{\textsc{Add Desc.}}\\
\multicolumn{2}{c|}{} & Train & Test & Train & Test & Train & Test & Train & Test \\
\hline
\multirow{2}{*}{r = 2}& Overall & 59.50(1.26) & 55.3(0.53) & 30.62(4.59) & 32.02(3.20) & 56.74(0.17) & 51.02(1.21) & 46.67(1.50) & 40.81(2.47) \\
& Per-Video & 58.54(0.29) & 55.11(1.34) & 31.09(3.57) & 32.30(2.35) & 54.99(0.43) & 49.57(1.37) & 46.09(2.00) &  40.11(1.23) \\
\hline
\multirow{2}{*}{r = 8}& Overall & 61.20(2.04) & 57.97(1.14) & 56.58(2.21) & 52.17(0.43) &62.56(0.02)  & 56.64(0.20) & 56.17(4.02) &  52.28(4.02) \\
& Per-Video & 58.72(1.69) &  61.20(2.04) & 54.87(2.01) & 51.03(0.85) &  60.58(0.32) & 57.34(0.26) & 55.05(3.76) &   52.79(3.77)\\
\hline
\multirow{2}{*}{r = 32}& Overall & 64.74(0.72) & 60.74(1.29) & 57.19(2.12) & 55.01(1.99) & 64.57(0.23) & 60.63(0.92) & 67.05(1.46) &  63.11(0.03)\\
& Per-Video & 61.83(1.06) & 57.27(0.96) & 54.87(2.08) & 53.07(2.65) & 62.07(0.23) & 58.99(1.48) & 63.74(1.84) & 60.46(0.14) \\
\hline
\multirow{2}{*}{r = 128}& Overall & 63.19(4.57) &  59.73(2.39) & 66.75(0.84) & 63.89(0.86) & 61.42(2.98) & 58.18(0.003) & 61.27(1.49) &   60.64(0.49) \\
& Per-Video & 59.76(3.42) & 57.15(1.87) & 63.28(0.38) &  61.22(1.08) &  58.50(2.61) &  54.65(0.003) & 58.87(1.38) &  58.30(1.10)\\
\hline
\multirow{2}{*}{r = 512}& Overall & 64.79(0.32) & 59.51(1.05) & 65.86(0.005) &61.4(0.001) & 64.93(3.98) &59.86(4.40) & 66.15(3.48) &61.21(0.86)  \\
& Per-Video &  60.58(0.22) & 56.83(1.01) & 62.61(0.005) & 58.4(0.1) & 61.49(3.15) & 57.85(3.55) & 62.86(3.02)  & 60.11(2.57)\\
\hline
% % \cline{2-5}
% & \textsc{VidMAE} & $0.66$ & $0.62 \pm 0.07$ & $0.7$ & $0.64 \pm 0.07$\\
% % \cline{2-5}
% & \textsc{ImageBind} & $\mathbf{0.69}$ & $\mathbf{0.66 \pm 0.08}$ & $\mathbf{0.75}$ & $\mathbf{0.69 \pm 0.08}$\\
% \hline
% \multirow{3}{*}{ \centering  \rotatebox[origin=c]{0}{\specialcell{ \textsc{Decoder}\\(Zero-shot)}}} & \specialcell{\textsc{Qwen 2.5VL} (3B)} & $0.28$ & $0.28 \pm 0.05$ & $0.26$ & $0.26 \pm 0.02$ \\
% % \cline{2-5}
% & \specialcell{\textsc{Qwen 2.5VL} (7B)} & $0.47$& $0.46 \pm (0.1)$ & $0.46$ & $0.45 \pm 0.08$  \\
% & \specialcell{\textsc{Qwen 2.5VL} (32B)} & $0.54$& $0.53 \pm (0.07)$ & $0.52$ & $0.5 \pm 0.07$ \\
% \cline{1-6}
% \multirow{2}{*}{ \centering  \rotatebox[origin=c]{0}{\specialcell{ \textsc{Decoder} \\ (Finetuned)}}} & \specialcell{\textsc{Qwen 2.5VL} (7B)-1.6fps} & $0.62$ & $0.59 \pm 0.09$ & $0.71$ & $0.67 \pm 0.06$ \\
% & \specialcell{\textsc{Qwen 2.5VL} (7B) - 3.2fps} & $0.62$ & $0.59 \pm 0.07$ & $0.66$ & $0.62 \pm 0.07$ \\
\hline
\end{tabular}

\label{tab:ablation}
\end{footnotesize}
\end{table*}

This section provides a detailed ablation study of the finetuning of the decoder. There are several design parameters we consider for the study. 
\begin{itemize}
    \item[--] \textbf{LoRA rank} (r) - The performance of the model generalization heavily depends on the rank used for the LoRA finetuning \cite{lora}. We explore the performance of finetuning for different rank (r) = $[ 2, 8, 32, 128, 512]$. We fix $\alpha = 2\times r$ throughout our experiments (following \cite{lora}). 
    \item[--] \textbf{Greedy vs Not greedy Decoding  (temperature = $0 \rightarrow 1.0$)} - the temperature parameter controls the level of randomness in the output by adjusting the probability distribution of the next word for the VLM. A greedy decoding (low temperature ~ 0) is expected to make the model more deterministic and focused, choosing more probable words for consistent and factual outputs. A high temperature increases randomness, leading to more creative and diverse outputs by giving more weight to less likely words. While greedy decoding is known to be more factual for generation use-cases, here we analyze its effect for the predictive use-case.
    \item[--] \textbf{Additional Label description} -  While the quality of (instruction) finetuning heavily depends on optimized instruction, very little has been explored on the quality/details in the label. For a predictive task a typical label would only include the class labels (`toprock', `footwork', `powermove', `None'). We explore the effect of adding additional (move) description during finetuning, towards the factual accuracy of the model's test-time generation. We try two different cases,
    \begin{itemize}
        \item[$\bullet$] Add Description - in this case we prepend additional description of the move type (break down of the move). And add an instruction `Analyzing this video this is the Move :: $<$move$>$' (see Fig \ref{fig_ft_prompt}). The goal here is to provide additional reasoning capability for the model during training. 
        \item[$\bullet$] No Description - we do not provide the move description in the label and only use :-  `Analyzing this video this is the Move :: $<$move$>$' (see Fig \ref{fig_ft_prompt})
    \end{itemize}
\end{itemize}

\noindent A complete snapshot of the effect of the different parameters is provided in Table \ref{tab:ablation}. This table provides the average (standard deviation) of test/train accuracy for move prediction over 3 separate runs of finetuning. We provide a detailed break down of these parameters separately and derive several insights on the decoder finetuning. These insights should also hold for other high-speed, intermittent-style sports' video level action classifications.

\subsubsection{Effect of LoRA rank on generalization}
\begin{figure}
  \begin{center}
    \includegraphics[width=0.45\textwidth]{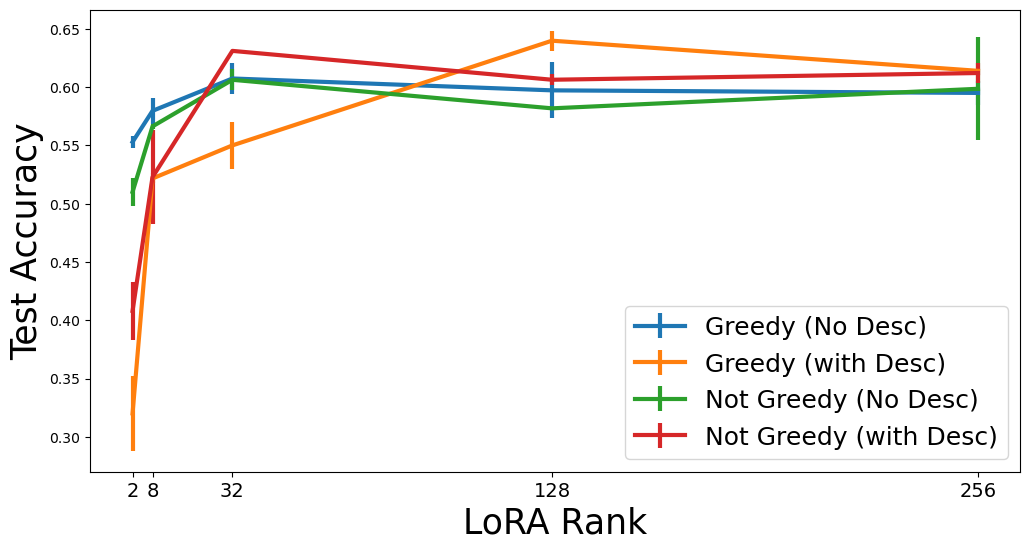}
  \end{center} 
  \caption{Variation in Test Accuracy for different LoRA ranks.} \label{fig_lora_rank} 
\end{figure} 

Fig \ref{fig_lora_rank} provides the variation in test accuracy for the different finetuned models for the different LoRA rank (r) = $[ 2, 8, 32, 128, 512]$. A clear trend seen here is that, for finetuning without additional description, the model performance has less variability. However, finetuning with additional description result to huge variations. Here, for a lower rank model, the generalization is severely degraded. This later improves (and stabilized) for higher rank (more - complex) models. This effect can be attributed to the typical underfitting phenomenon, which indicates that correctly generating more complex descriptions would need more complex models (i.e. higher rank). In addition, well known from prior literature \cite{lora}, proper tuning of the LoRA rank hyper-parameters are of utmost importance for accurate generation (even for predictive tasks). 
\subsubsection{Greedy vs Not Greedy Decoding}

\begin{figure}
  \centering
  \begin{subfigure}{0.45\linewidth}
    \includegraphics[width=1.\textwidth]{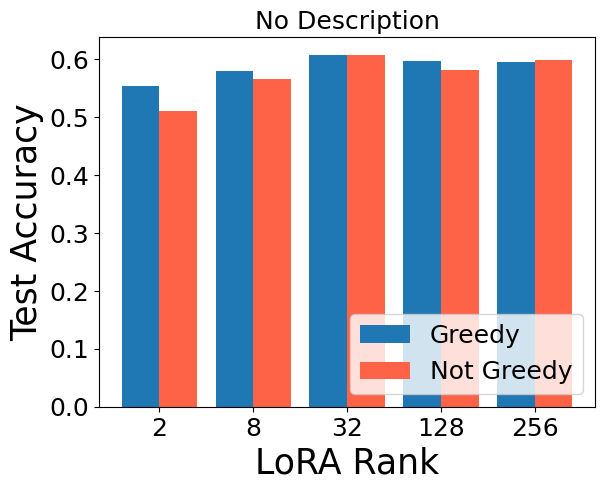}
    \caption{No Description}
    \label{fig:no_desc_gvng}
  \end{subfigure}
  \begin{subfigure}{0.45\linewidth}
    \includegraphics[width=1.\textwidth]{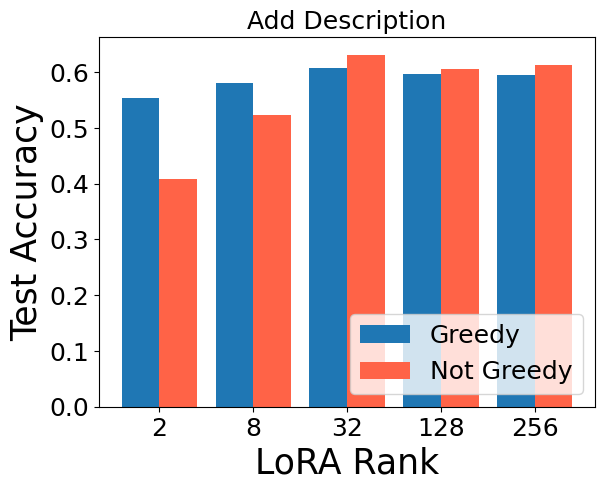}
    \caption{Add Description}
    \label{fig:add_desc_gvng}
  \end{subfigure}
  \caption{Comparison of Test Accuracy for Greedy (Temperature  $T = 0.0$) vs not greedy (Temperature  $T = 1.0$) decoding.}
  \label{fig:greedyvsnotgreedy}
\end{figure}

Figure \ref{fig:greedyvsnotgreedy} provides a focused analysis of the effect of Temperature $T = 0 \rightarrow 1$. We show the results of the effect of greedy ($T = 0$) vs not greedy ($T=1$) decoding for both no description (see Fig. \ref{fig:no_desc_gvng})and additional description setups (see Fig. \ref{fig:add_desc_gvng}). For both these cases greedy decoding performs similarly over all ranks. However, while not greedy version performs poorly for low rank models (less complex models), they provide significant improvements for more complex setups (higher rank models). Of course, beyond particular complexity its stagnates out. Note that, while greedy decoding prefers higher probability tokens as compared to higher temperature $T = 1.0$, that does not necessarily translate to more factual accuracy! Overall, the results indicate that a non-greedy model with sufficient model complexity (rank) is preferred over greedy decoding for better predictive performance.

\subsubsection{Effect of Additional Label Descriptions}

\begin{figure}
  \centering
  \begin{subfigure}{0.45\linewidth}
    \includegraphics[width=1.\textwidth]{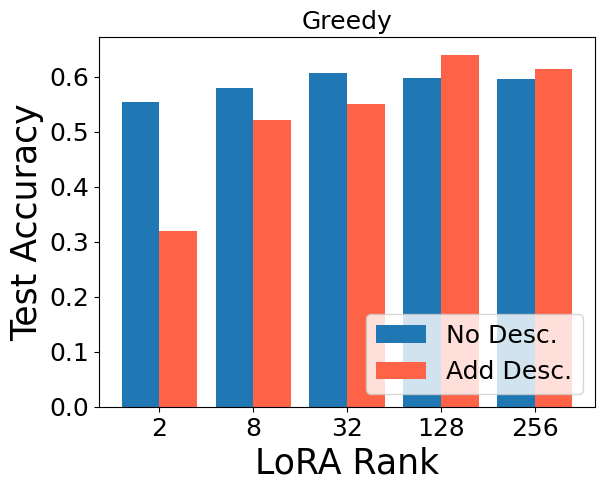}
    \caption{Greedy $T = 0$}
    \label{fig:greedy_desc}
  \end{subfigure}
  \begin{subfigure}{0.45\linewidth}
    \includegraphics[width=1.\textwidth]{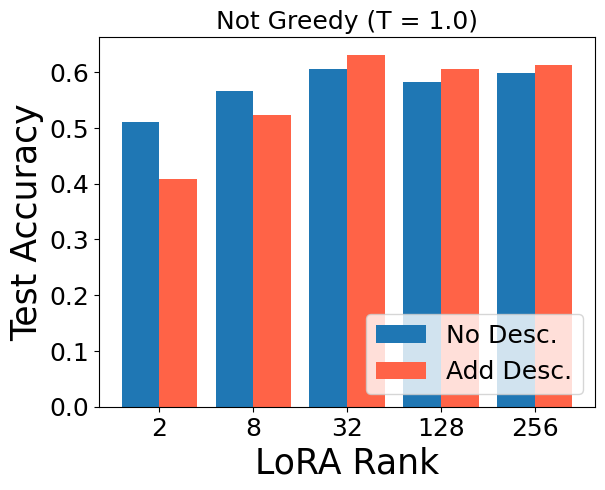}
    \caption{Not Greedy $T=1.0$}
    \label{fig:notgreedy_desc}
  \end{subfigure}
  \caption{Comparison of Test Accuracy for No description vs Additional description in label during finetuning.}
  \label{fig:add_vs_no_desc}
\end{figure}

We see very similar properties (as the non-greedy version) for the added label description. It seems for low rank (complexity) setups having the additional description results to poor generalization. However, for a sufficiently complex (rank) model, an additional description in the label provides an improved generalization. 

In essence, the effect of having a higher temperature $T>0$ or additional label description, behaves as adding noise in the decision probability space or label smoothing respectively. This overall acts as a regularization parameter (and contributes to model bias). For a sufficient LoRA rank, we can avoid underestimation and achieve improved generalization.
\section{Conclusion}\label{sec:conclusion}
This work provides a detailed comparison of the existing state-of-the-art Encoder vs Decoder based video models for breakdance moves classification. Even though there have been recent results of decoder based models (LLMs) improving over traditional encoder based models for text classification; this does not generalize to high-speed sports like breakdance video classification. In addition, we share some simple yet effective mechanism to determine the effectiveness of the different encoder (embedding) models using a linear discriminant analysis. Our results show encoder models targeted towards capturing the generic semantic understanding (like ImageBind) are more amenable for task transfers compared to ViViT or VidMAE pre-trained for specific tasks (kinetics-400). Finally, we provide a thorough ablation study of the decoder-based finetuning. Our analysis provides several insights towards improving the predictive performance of generative models. Our analysis shows non-greedy decoding is preferred to greedy decoding for factual accuracy! Moreover, adding additional descriptions to the labels could improve the overall accuracy. In essence, such techniques may provide regularization through noise, and lead to improve predictive performance for Video level LLM generators.

{
    \small
    \bibliographystyle{ieeenat_fullname}
    \bibliography{main}
}

% WARNING: do not forget to delete the supplementary pages from your submission 
\clearpage
\maketitlesupplementary

\appendix

\section{BRACE Dataset - Additional Details}
\label{sec:app_brace_details}

\begin{figure}
  \begin{center}
  \vspace{-0.2cm}
    \includegraphics[width=0.25\textwidth]{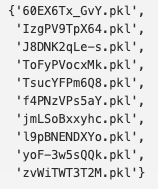}
  \end{center} 
  \caption{Brace Test Videos.} \label{fig_test_brace} 
\end{figure}

Fig \ref{fig_test_brace} provides the list of 10 videos we use as test for reproducibility. The remaining 71 videos are used for training.

\end{document}